\documentclass{article} 

\usepackage[preprint]{colm2026conference}
\usepackage{microtype}
\usepackage{hyperref}
\usepackage{url}
\usepackage{booktabs}
\usepackage{lineno}
\usepackage{makecell}
\usepackage{tikz}
\usepackage{amsmath,amssymb,amsfonts}
\usepackage{algorithmic}
\usepackage{graphicx}
\usepackage{textcomp}
\usepackage{soul}
\usepackage{xcolor}
\usepackage{placeins}

\usepackage{amsmath,amsfonts,bm}

\def\eqref#1{equation~\ref{#1}}

\def\1{\bm{1}}

\DeclareMathAlphabet{\mathsfit}{\encodingdefault}{\sfdefault}{m}{sl}
\SetMathAlphabet{\mathsfit}{bold}{\encodingdefault}{\sfdefault}{bx}{n}

\usepackage{graphicx}
\usepackage{url}
\usepackage{color}
\usepackage{alltt}
\usepackage{syntax}
\usepackage{newfloat}
\usepackage{multicol}
\usepackage{amsmath}
\usepackage{tikz}
\usepackage{filecontents} 
\usepackage{graphicx}
\usepackage{epstopdf} 
\usepackage{xspace}
\usepackage{enumitem}
\usepackage{svg}
\usepackage[ruled,linesnumbered,noend]{algorithm2e}
\SetKw{Continue}{continue}
\usepackage{adjustbox}
\usepackage{wrapfig}
\usepackage{graphbox}
\usepackage{rotating}
\usepackage{pdfpages}
\usepackage{mdwlist}
\usepackage{multirow}
\usepackage{booktabs}
\usepackage{framed}

\usepackage{subcaption}
\usepackage[labelfont=bf,skip=0pt]{caption}
\usepackage{hyperref} 
\hypersetup{
	citebordercolor={0 1 0},
	citecolor=blue,
	colorlinks=false,
	filebordercolor={0 .5 .5},
	filecolor=green,
	linkbordercolor={1 0 0},
	linkcolor=blue,
	menubordercolor={1 0 0},
	pageanchor=true,
	pagebackref=true,
	pdfborder={0 0 1},
	pdfpagelabels=true,
	pdftex,
	plainpages=false,
	urlbordercolor={0 1 1},
	urlcolor=blue,
}

\usepackage[capitalize,nameinlink]{cleveref}
\hypersetup{%
	bookmarksnumbered, bookmarksopen=true, bookmarksopenlevel=1,%
}
\crefname{figure}{Figure}{Figures}
\crefname{listing}{Query}{Queries}
\crefname{section}{Section}{Sections}
\crefname{table}{Table}{Tables}
\crefname{BNF}{Grammar}{Grammars}
\crefname{algorithm}{Algorithm}{Algorithms}
\crefname{equation}{Equation}{Equations}

\usepackage{listings}
\definecolor{mygreen}{rgb}{0,0.6,0}
\definecolor{mygray}{rgb}{0.5,0.5,0.5}

\DeclareFloatingEnvironment[
fileext   = logr,
listname  = {List of Grammars},
name      = Grammar,
placement = !htbp
]{BNF}

\newcommand{\msim}{\raise.17ex\hbox{$\scriptstyle\sim$}}

\newcommand{\eat}[1]{}

\newcommand{\tool}{\textsc{GRID}\xspace}
\newcommand{\toolfull}{Graph Representation of Intelligence Data (\textbf{G.R.I.D.})\xspace}

\usepackage{pifont}

\usepackage{tikz}

\definecolor{darkblue}{rgb}{0, 0, 0.5}
\hypersetup{colorlinks=true, citecolor=darkblue, linkcolor=darkblue, urlcolor=darkblue}

\begin{document}

\ifcolmsubmission
\linenumbers
\fi

\title{GRID: Graph Representation of Intelligence Data for Security Text Knowledge Graph Construction}

\author{
Liangyi Huang$^{1}$ \quad Zichen Liu$^{1}$ \quad Fei Shao$^{2}$ \quad Shang Ma$^{3}$ \\
Mengshi Zhang$^{4}$ \quad Zihao Chen$^{5}$ \quad Yanfang Ye$^{3}$ \quad Xusheng Xiao$^{1}$ \\
$^{1}$Arizona State University \quad $^{2}$Case Western Reserve University \quad $^{3}$University of Notre Dame \\
$^{4}$TensorBlock \quad $^{5}$Facebook \\
\texttt{lhuan139@asu.edu}
}
\maketitle
\begin{abstract}
Security knowledge graphs can serve as computable and traceable external memory for security agents. Our goal is to equip LLMs with security-domain knowledge for knowledge graph extraction from long-form security text. However, existing LLMs largely lack such domain knowledge grounded in real security text, and end-to-end document-to-graph training is difficult to supervise with cheap and stable rewards. We present \toolfull, an end-to-end framework for security text knowledge graph construction. \tool first constructs security-domain supervision from security-related CTI articles in an unsupervised manner by constructing traceable article-graph alignments through graph extraction and knowledge-graph-conditioned text revision. It then reformulates document-to-graph learning into a scripted task bank that combines four-option multi-select questions with triple-level regex matching targets, yielding cheaper and more stable task-specific rewards than asking an LLM judge to score full graph outputs at every training step. Based on this supervision pipeline, we train two Qwen3-4B-Instruct-2507-based 4B extractors: a primary Task-bank Reward model and a secondary End2End Reward model for direct article-to-knowledge-graph generation with LLM-as-judge precision/recall rewards. On a unified benchmark of 249 CTI articles from five sources: GRID, CASIE, CTINexus, MalKG, and SecureNLP, the post-trained Task-bank Reward model together with the ontology-guided GRID extraction pipeline reaches 84.62\% source-averaged precision, 64.91\% source-averaged recall, and 68.53\% Avg F1, achieving the best source-averaged recall and a near-tied top Avg F1 with much less token usage and lower deployment cost. The secondary End2End Reward model reaches 76.91\% source-averaged precision, 53.85\% source-averaged recall, and 58.06\% Avg F1. Further analyses show that the task-bank reward can be constructed once offline and reused across later post-training runs while outperforming the online End2End LLM-as-judge reward as well as weaker alternatives such as Choice-only Reward and End2End SFT without RL, and that both article rewriting and article-complexity-ordered training are necessary for the best performance.


\end{abstract}

\section{Introduction}
\label{sec:intro}
In recent years, cyber attacks have become more frequent, more complex, and more expensive \cite{gao2018aiql,apt,depimpact,target}. The 2024 Report on the Cybersecurity Posture of the United States shows that reported ransomware incidents increased by 22\% since 2022, and the related costs increased by 74\% \cite{cybersecurityposture2024}. However, many organizations still lack a good understanding of current threats. A recent study shows that 79\% of security decision-makers often ignore threat actor information, only 35\% believe their organizations understand adversaries' tactics, techniques, and procedures, and 68\% think their threat intelligence capabilities need major improvement \cite{mandiant2024global}.

Cyber Threat Intelligence (CTI) is therefore important for cyber defense \cite{cti,cti2}. Common structured CTI sources, such as Indicators of Compromise (IOCs), Common Vulnerabilities and Exposures (CVEs), and cyber kill chains, are useful but often miss important attack context \cite{ioc,ioc2,ioc3,ctifeed1,ctifeed2,cve,cyberkillchain,mitreattack}. In contrast, unstructured CTI articles, such as technical blogs and threat reports, often describe attack behaviors, attacker goals, exploited vulnerabilities, and malware evolution in much more detail \cite{ioc2,dong2019towards}. This has motivated efforts to organize CTI knowledge into structured resources such as MITRE ATT\&CK and NVD, and to automatically extract threat knowledge from CTI articles \cite{mitreattack,nvd,li2022attackg,satvat2021extractor,gao2022threatkg,llms,chatgpt,huang2024ctikg}.

\begin{figure*}[t]
\centering
\includegraphics[width=0.95\textwidth]{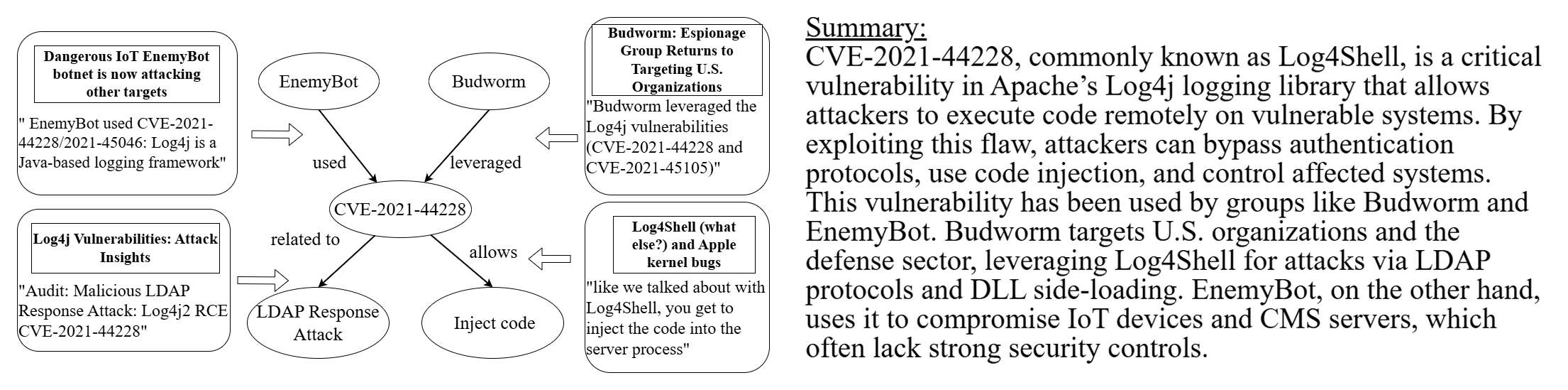}
\caption{Log4Shell (CVE-2021-44228) CTI articles: knowledge graph (left) and attack summary (right)}
\label{fig:exampleintro}
\end{figure*}

Knowledge graphs are useful for this purpose because they represent both security entities and their relations in one structure. Prior work has shown that graph-based threat knowledge helps forensic analysis and attack reconstruction by matching known threat behaviors with system auditing events and provenance graphs \cite{poirot,backtracking,depimpact,depcomm}. More generally, graph-structured knowledge also helps LLMs and agents reason over connected evidence. Think-on-Graph reports state-of-the-art results on 6 of 9 reasoning benchmarks, G-Retriever improves valid-node grounding from 31\% to 77\% and fully valid graph grounding from 8\% to 62\%, and recent graph-memory systems for agents report a 26\% relative improvement over memory baselines and up to 18.5\% higher accuracy on long-horizon tasks \cite{sun2023think,he2024g,chhikara2025mem0,rasmussen2501zep}.
Figure~\ref{fig:exampleintro} shows a Log4Shell example with a knowledge graph and its attack summary from CTI articles.

In this paper, we propose \toolfull, a low-cost framework for security text knowledge graph construction.
Rather than treating security knowledge graph extraction as a standalone prompting problem, \tool provides a complete pipeline for developing and assessing LLM-based security knowledge graph extraction systems.
This pipeline spans unsupervised supervision construction from CTI articles, cheaper task-bank rewards for end-to-end document-to-graph learning, a fixed two-prompt inference pipeline, and trustworthy automatic evaluation against human-annotated CTI data.

\noindent\textbf{Challenges}.
We next summarize the key challenges faced by existing approaches. 

\begin{itemize}[noitemsep, topsep=1pt, partopsep=1pt, listparindent=\parindent, leftmargin=*]
\item \textit{Lack of Integration of Security Domain Knowledge}:
Most existing LLMs are designed for general-purpose use rather than knowledge graph extraction from security text, especially at small model scales, leaving practitioners dependent on expensive commercial APIs.

\item \textit{Lack of High-Quality CTI Article-Graph Alignment Data}:
Training LLMs for CTI knowledge graph extraction requires supervision that tightly aligns real CTI text with graph outputs, but there is currently no public high-quality dataset that provides such article-graph pairs for end-to-end training.

\item \textit{Expensive Reward Signals for Open-Ended Knowledge Graph Extraction}:
End-to-end knowledge graph extraction is an open-ended generation task, making reward design for reinforcement learning difficult and expensive.
Even when LLM-as-judge is available, asking it to score full extracted graphs at training time still incurs high cost.

\item \textit{Shallow Shortcut Learning in Relation Extraction}:
LLMs tend to rely on superficial lexical overlap, local co-occurrence, or other surface heuristics when predicting relations, instead of deeply understanding entity semantics, aliases, structural hierarchy, and relation constraints in CTI narratives.
\end{itemize}

\noindent\textbf{Contributions}.
\begin{itemize}[noitemsep, topsep=0pt, partopsep=0pt, listparindent=\parindent, leftmargin=*]

\item \textit{Automatic Annotation of Article-Graph Alignment Data}:  
\tool introduces an automatic data annotation algorithm for CTI knowledge graph extraction.
It first generates a traceable knowledge graph that preserves verbatim evidence anchors from the source text, and then performs knowledge-graph-conditioned text revision to remove CTI information that is not captured by the graph while retaining non-CTI context.
This yields high-quality article-graph alignments without requiring large-scale manual annotation.

\item \textit{Low-Cost Task-Bank Reformulation for RL Training}:  
\tool reformulates open-ended knowledge graph extraction into scripted supervision tasks that combine four-option multi-select questions with triple-level regex targets.
This replaces full-graph scoring with cheaper task-level checks.

\item \textit{Ontology-Guided CTI Knowledge Graph Extraction}:  
\tool designs a CTI-oriented ontology that explicitly models entity types, relation categories, aliases, and hierarchy, so that extraction depends on entity semantics and constraints rather than only shallow textual cues; together with the post-trained model, this lowers deployment cost.

\item \textit{Out-of-the-box Benchmark and Trustworthy Automatic Evaluation}:  
To address the lack of an out-of-the-box benchmark for CTI knowledge graph extraction, \tool also builds a test benchmark centered on real CTI articles.
The benchmark combines manually annotated real-world CTI data with multiple existing security text datasets, and is paired with a trustworthy automatic evaluator based on text-provable precision and recall.
\end{itemize}

\noindent\textbf{Evaluations}.  
We evaluate \tool on a unified benchmark of 249 CTI articles from five sources, comprising 49 GRID articles (avg.\ 1,102 tokens and 15.35 ground-truth edges), 50 CASIE articles (avg.\ 537 tokens and 7.94 ground-truth edges), 50 CTINexus articles (avg.\ 191 tokens and 11.80 edges), 50 MalKG articles (avg.\ 6,632 tokens and 48.90 edges), and 50 SecureNLP articles (avg.\ 11,000 tokens and 68.66 edges), after removing articles whose ground-truth knowledge graphs contain fewer than five edges. For all RQs, we report effectiveness results using a calibrated LLM judge that reaches 86.0\% agreement with annotations from three human reviewers. Using Qwen3-4B-Instruct-2507, we train two 4B extractors: a primary Task-bank Reward model and a comparison model trained with online End2End LLM-as-judge reward. On this benchmark, the post-trained Task-bank Reward model together with the ontology-guided GRID pipeline achieves 84.62\% source-averaged precision, 64.91\% source-averaged recall, and 68.53\% Avg F1, giving the best source-averaged recall and a near-tied top Avg F1 with much less token usage than CTINexus. The online End2End LLM-as-judge reward model reaches 76.91\% precision, 53.85\% recall, and 58.06\% Avg F1. RQ2 ablations further validate Task-bank Reward as an effective reward design, outperforming online End2End reward, Choice-only Reward, End2End SFT without RL, and the base model. Under the same training budget, \tool's full setting achieves higher training reward and a higher test-set score reflecting precision and recall than the variants without article rewriting or article-complexity ordering. Code and data are accessible at \url{https://github.com/anonymousauthorname/ProjectGRID}~\cite{gridrepo}.

\section{Approach}
\label{sec:approach}
\begin{figure*}[t]
\centering
\includegraphics[width=0.98\textwidth]{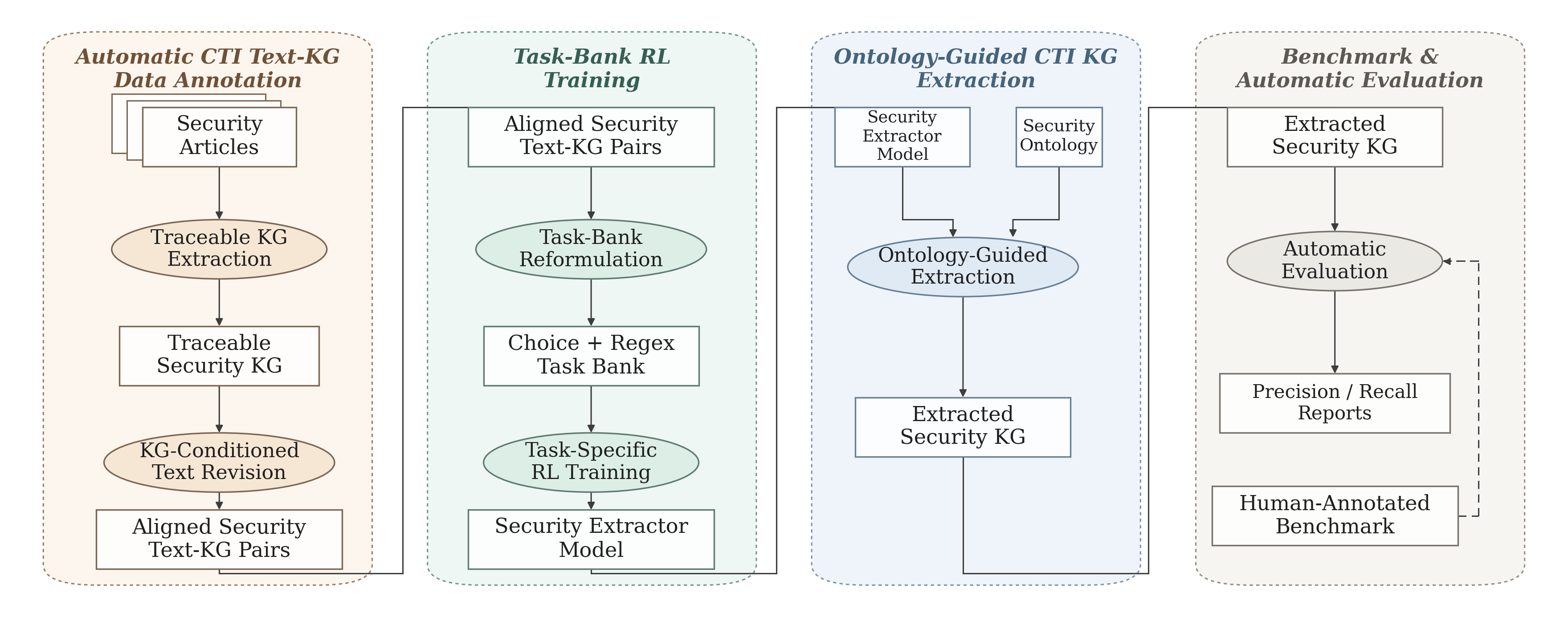}
\caption{Overview of \tool}
\label{fig:architecture}
\end{figure*}

Figure~\ref{fig:architecture} summarizes the overall pipeline of \tool, and the rest of this section explains each step in turn.

\subsection{Automatic Annotation of Article-Graph Alignment Data}
\tool maps each raw CTI article to an aligned pair $(a', G')$ through a two-stage annotate-and-revise loop. It first extracts a traceable knowledge graph under a strict text-provable constraint, and then rewrites the article against that graph so that unsupported security content is removed while graph-grounded evidence and non-security context are preserved. The result is a revised article whose security-bearing content is explicitly aligned with the extracted graph; Algorithm~\ref{alg:grid-data-annotation} gives the procedure.

\begin{algorithm}[t]
\caption{Automatic Annotation of Article-Graph Alignment Data}
\label{alg:grid-data-annotation}
\DontPrintSemicolon
\KwIn{Raw CTI article $a$, traceable extraction prompt $P_{\mathrm{trace}}$, revision prompt $P_{\mathrm{rev}}$}
\KwOut{Article-graph alignment $(a', G')$ consisting of revised article $a'$ and text-grounded knowledge graph $G'$}

$G \leftarrow \mathrm{LLMExtract}(a, P_{\mathrm{trace}})$\;
Parse $G$ into entity list $E$ and relation list $R$\;
\ForEach{$r \in R$}{
    keep sentence-local subject/object mentions in $(r.\mathrm{sub}, r.\mathrm{obj})$\;
    keep verbatim evidence anchors in $(r.\mathrm{raw\_sub\_name}, r.\mathrm{raw\_obj\_name}, r.\mathrm{raw\_text\_start}, r.\mathrm{raw\_text\_end})$\;
}
Mark all anchor spans in $a$ that are protected by $R$\;
$a' \leftarrow \mathrm{LLMRevise}(a, G, P_{\mathrm{rev}})$, deleting unsupported security mentions while keeping protected anchors and non-security context\;
$G' \leftarrow (E, R)$\;
\Return $(a', G')$\;
\end{algorithm}

\subsection{Task Bank Construction}
Rather than directly rewarding full document-to-graph generation, \tool converts each article-graph alignment $(a', G')$ into two easy-to-check RL task families. The first creates four-option multi-select questions, where the ground-truth answer can contain any subset of 0--4 correct options. The second creates one triple-level regex target for each ground-truth KG edge, so that graph supervision can be reduced to per-edge matching rather than whole-graph judging.

On the choice side, distractors are contrastive negatives that may look related in the article but are actually invalid, or may seem reasonable based on real-world CTI experience even though the article itself does not support them. On the regex side, matching is normalized at the entity and relation levels but remains edge-aligned. Table~\ref{tab:step5a-submodes} and Table~\ref{tab:regex-rules} summarize the two rule families.

\begin{table}[t]
\centering
\setlength{\tabcolsep}{3pt}
\footnotesize
\caption{Choice-question patterns for precision and hallucination checks}
\label{tab:step5a-submodes}
\begin{tabular}{p{0.16\linewidth}p{0.24\linewidth}p{0.54\linewidth}}
\toprule
\textbf{Family} & \textbf{Pattern} & \textbf{Definition} \\
\midrule
\multirow{2}{*}{Precision} &
Supported triples &
Among the four options, 0-4 are text-supported; the rest are near-miss distractors with a wrong subject, object, or relation. \\

&
Incorrect triples &
Among the four options, 0-4 are deliberately incorrect; the rest are supported triples. \\

\midrule
\multirow{5}{*}{Hallucination} &
Relation illusion &
The subject and object are grounded, but their relation is unsupported. \\

&
Object illusion &
The subject and relation are grounded, but the object is unsupported. \\

&
Subject illusion &
The relation and object are grounded, but the subject is unsupported. \\

&
Total illusion &
All three triple elements are unsupported, but the triple remains CTI-plausible. \\

&
Partial illusion &
Only one triple element is grounded; the other two are unsupported. \\
\bottomrule
\end{tabular}
\end{table}

\begin{table}[t]
\centering
\setlength{\tabcolsep}{3pt}
\footnotesize
\caption{Regex rules for triple-level matching}
\label{tab:regex-rules}
\begin{tabular}{p{0.22\linewidth}p{0.70\linewidth}}
\toprule
\textbf{Aspect} & \textbf{Rule} \\
\midrule
Edge alignment &
Maintains one regex triplet per ground-truth edge. \\

Entity normalization &
Normalizes aliases, abbreviations, number, hyphenation, and shortened head nouns for subjects and objects. \\

Role-preserving generalization &
Permits limited generalization only under role-preserving equivalence. \\

Relation normalization &
Normalizes inflection, voice, prepositions, and common CTI paraphrases. \\
\bottomrule
\end{tabular}
\end{table}

\subsection{Article Complexity Modeling}
\tool assigns each source article a scalar complexity score so that the exported training set can be reordered from easy to hard. In the current training pipeline, complexity is defined only at the article level, and all reward-related prompts derived from the same article inherit the same score.

For an article $a$, \tool computes
\[
C_{\mathrm{article}}(a)=\frac{1}{2}C_{\mathrm{base}}(a)+\frac{1}{2}C_{\mathrm{graph}}(a),
\]
where $C_{\mathrm{base}}$ averages percentile ranks of article length, entity count, relation count, and text-normalized entity/relation density, while $C_{\mathrm{graph}}$ averages percentile ranks of alias, connectivity, span, and fixed-order crossing statistics.
This article-level score is attached to every training item derived from the same article and is later used for article-complexity-ordered RL training.

\subsection{Reward Design for RL Training}
The RL stage trains the LLM model with Soft Adaptive Policy Optimization (SAPO), an improved GRPO-style variant~\cite{gao2025soft}. In the current implementation, task-bank rewards are computed locally by task type. Both task types additionally receive a 0.1 format reward when the output follows the required reasoning-before-answer format. For a choice task, the main reward is 1.0 when the predicted answer set exactly matches the ground-truth answer set, 0.5 when the two sets overlap but are not identical, and 0.0 otherwise.

For a regex task, the main reward is $r_{\mathrm{regex}}=n_{\mathrm{match}}/n_{\mathrm{gt}}$, where $n_{\mathrm{match}}$ is the number of matched ground-truth regex triplets, $n_{\mathrm{gt}}$ is the total number of ground-truth regex triplets, and a ground-truth regex triplet is counted as matched only when its subject, relation, and object are all matched by the same predicted triple.

\subsection{Ontology-Guided Security Knowledge Graph Extraction}
The ontology is not used as a post-hoc label list. Instead, it guides the LLM to recover entity relations that are not directly stated through explicit verbs or relative clauses, by providing typed entity categories, normalized relation families, aliases, and hierarchy constraints. Appendix Figure~\ref{fig:ontology-prompt-block} gives a prompt-block view, and the full ontology inventory, including predefined entity and relation types, is summarized in Appendix~\ref{app:ontology}.

It enforces text-provable truth by forbidding external completion, subject elevation, behavior-to-structure conversion, and subject-changing chain deduction; treats alias, lineage, categorization, and normalized relation types as ontology-native semantics rather than post-hoc cleanup; and uses a connectivity recheck to remove unsupported isolated entities instead of hallucinating links. In GRID inference, Step 1 proposes candidate entities, and Step 2 finalizes entities and relations against the original article.

\subsection{LLM-Based Automatic Evaluation}
\label{sec:judge-protocol}
For end-to-end knowledge graph extraction, \tool uses an LLM-based evaluator that scores precision and recall with two prompt templates governed by the same text-provable principle as the extractor. The precision prompt audits predicted edges one by one and asks whether each predicted edge is either directly supported by the article text or equivalent to a ground-truth edge under the judge rules. The recall prompt audits ground-truth edges one by one and asks whether each ground-truth edge is captured by the predicted graph, either by a directly equivalent predicted edge or by a text-supported, subject-preserving combination of predicted edges allowed by the judge rules. In both prompts, the judge is required to return the edge index being audited, a binary verdict, and either supporting text evidence or the matched counterpart predicted edge(s). The full judge rules are summarized in Appendix~\ref{app:judge-rules}, Table~\ref{tab:judge-rules}.

\section{Evaluation}
\label{sec:evaluation}

In the evaluations, we focus on the following research questions:
\begin{itemize}[noitemsep, topsep=2pt, partopsep=1pt, listparindent=\parindent, leftmargin=*]
    \item RQ1: How do the two main \tool systems compare with representative CTI knowledge-graph baselines?
    \item RQ2: How do the main post-training designs differ in effectiveness and engineering cost?
    \item RQ3: How much do article rewriting and article-complexity ordering matter within the primary Task-bank Reward setup?
\end{itemize}

\subsection{Evaluation Setup}

\noindent\textbf{Evaluation Dataset}.
Our benchmark contains 249 CTI articles drawn from five sources after removing articles whose ground-truth knowledge graphs contain fewer than five edges. We first manually annotate 59 real-world CTI articles and retain 49 filtered GRID articles (avg.\ 1,102 tokens and 15.35 edges). We then apply the same ground-truth-graph-size filter to four external sources and retain 50 articles each from CASIE (537 tokens and 7.94 ground-truth edges per article on average), CTINexus (avg.\ 191 tokens and 11.80 edges), MalKG (avg.\ 6,632 tokens and 48.90 edges), and SecureNLP (avg.\ 11,000 tokens and 68.66 edges)~\cite{satyapanich2020casie,cheng2025ctinexus,rastogi2021information,phandi2018securenlp}.

\noindent\textbf{Implementation}.
We implement training with VERL~\cite{sheng2024hybridflow} and adopt Qwen3-4B-Instruct-2507~\cite{qwen3tech} as the base extractor model. This 4B backbone is strong enough for post-training while still keeping repeated SFT/RL ablations affordable. The rationale for this backbone choice is further discussed in Section~\ref{sec:discussion}. Additional implementation details are deferred to Appendix~\ref{app:training-details}.

\noindent\textbf{Training Setup}.
For the two main RQ1 systems, the primary Task-bank Reward model is trained on 800 CTI articles, whereas the secondary End2End Reward model is trained on 1,000 CTI articles. Task-bank Reward and End2End Reward use training batch sizes of 16 and 64, respectively, with rollout counts of 4 and 8. During training-data construction and the End2End training-side reward computation, we use GPT-5.3 Codex Medium~\cite{openai2026gpt53codex}. At test time, all models receive the full article and are evaluated on full-document extraction. Additional training and testing details are summarized in Appendix~\ref{app:training-details}.

\noindent\textbf{LLM Judge Calibration}.
Because exact string matching would unfairly penalize many semantically correct but surface-divergent extractions, we adopt the indexed edge-wise evaluation protocol described in Section~\ref{sec:judge-protocol}. We first fixed the GPT-5.4 mini judge configuration, with reasoning effort set to medium, temperature 0.1, and a maximum output budget of 65,536 tokens. We then manually inspected pilot judge outputs and iteratively refined the prompts by adding more detailed decision rules and representative examples. After freezing the final prompt set, we calibrated the judge against human judgments on 378 manually reviewed audit items from three human reviewers. The resulting agreement is 80.6\% for precision (154/191) and 91.4\% for recall (171/187), yielding an overall agreement of 86.0\% (325/378) with human annotations; detailed calibration slices and confusion matrices are deferred to Appendix~\ref{app:judge-calibration}. The same final judge configuration and prompt set were then used for all RQs.

\noindent\textbf{Baselines}.
For RQ1, we compare two \tool systems with eight representative baselines. The two \tool models are the primary Task-bank Reward system and the secondary End2End Reward system. The eight external baselines are CTINexus~\cite{cheng2025ctinexus}, CTIKG~\cite{huang2024ctikg}, Cognee~\cite{cogneerepo}, LLM-CAKG~\cite{wang2025automated}, Graphiti~\cite{graphitirepo}, GraphRAG~\cite{graphragrepo}, KnowGL~\cite{rossiello2023knowgl}, and AttacKG+~\cite{zhang2025attackg+}. All LLM-based systems are instantiated with the same base model as \tool, namely Qwen3-4B-Instruct-2507~\cite{qwen3tech}, with temperature 0.7 and a maximum generation budget of 32,768 tokens. We additionally surveyed LLM-TIKG~\cite{hu2024llm}, CTI-Thinker~\cite{yang2026cti}, CodeKGC~\cite{bi2024codekgc}, SecTKG~\cite{sun2023sectkg}, and the related resources CS13K~\cite{li2024cybersecurity}, HRTC~\cite{yue2024hrtc}, and BVTED~\cite{liu2024bvted}, but did not include them in this work because of incomplete public implementations, unavailable source data, or access restrictions. Among non-LLM methods, KnowGL is the only one retained in the main table; REBEL~\cite{huguet-cabot-navigli-2021-rebel-relation} and EXTRACTOR~\cite{satvat2021extractor} are not used because their performance is clearly lower.

\subsection{RQ1: Comparison with Baselines}
\begin{table*}[t]
\centering
\footnotesize
\setlength{\tabcolsep}{2.8pt}
\caption{Per-source precision, recall, and F1 (\%). Avg columns are arithmetic means over the five sources}
\label{tab:rq1-main}
\resizebox{\textwidth}{!}{
\begin{tabular}{lcccccccccccccccccc}
\toprule
\textbf{Method} & \multicolumn{15}{c}{\textbf{Source}} & \multicolumn{3}{c}{\textbf{Avg}} \\
\cmidrule(lr){2-16} \cmidrule(lr){17-19}
 & \multicolumn{3}{c}{\textbf{CASIE}} & \multicolumn{3}{c}{\textbf{CTINexus}} & \multicolumn{3}{c}{\textbf{GRID}} & \multicolumn{3}{c}{\textbf{MalKG}} & \multicolumn{3}{c}{\textbf{SecureNLP}} & \multicolumn{3}{c}{} \\
\cmidrule(lr){2-4} \cmidrule(lr){5-7} \cmidrule(lr){8-10} \cmidrule(lr){11-13} \cmidrule(lr){14-16} \cmidrule(lr){17-19}
 & P & R & F1 & P & R & F1 & P & R & F1 & P & R & F1 & P & R & F1 & P & R & F1 \\
\midrule
\tool (Task-bank) & 81.80 & \textbf{77.68} & \textbf{77.22} & 86.69 & 81.10 & 82.58 & 84.18 & \textbf{78.11} & \textbf{79.62} & 84.43 & 38.55 & 44.66 & \textbf{85.97} & 49.10 & 58.58 & 84.62 & \textbf{64.91} & 68.53 \\
\tool (End2End) & 80.88 & 66.98 & 70.76 & 80.30 & 76.67 & 76.83 & 78.71 & 64.93 & 67.35 & 74.94 & 24.87 & 31.61 & 69.71 & 35.81 & 43.77 & 76.91 & 53.85 & 58.06 \\
CTINexus & 83.44 & 61.41 & 67.49 & 86.75 & \textbf{91.02} & \textbf{87.83} & 83.62 & 71.64 & 75.91 & \textbf{88.71} & \textbf{39.42} & \textbf{48.24} & 85.07 & 53.88 & 63.80 & 85.52 & 63.47 & \textbf{68.66} \\
Cognee & 48.87 & 43.75 & 39.79 & 63.98 & 64.58 & 61.80 & 57.13 & 52.01 & 50.24 & 63.74 & 27.73 & 30.33 & 68.04 & 48.74 & 52.54 & 60.35 & 47.36 & 46.94 \\
LLM-CAKG & 85.39 & 52.76 & 58.81 & 79.90 & 55.63 & 63.41 & 80.65 & 69.61 & 72.60 & 78.49 & 38.27 & 46.70 & 80.35 & \textbf{65.16} & \textbf{68.70} & 80.96 & 56.29 & 62.04 \\
Graphiti & 70.50 & 24.39 & 32.40 & 70.87 & 35.05 & 44.39 & 69.97 & 46.13 & 51.50 & 74.12 & 32.14 & 39.02 & 71.78 & 33.88 & 43.35 & 71.45 & 34.32 & 42.13 \\
CTIKG & 87.41 & 34.51 & 44.57 & 82.67 & 40.82 & 50.25 & \textbf{84.28} & 43.06 & 52.30 & 79.20 & 18.91 & 26.29 & 84.12 & 40.61 & 51.12 & 83.54 & 35.58 & 44.91 \\
GraphRAG & \textbf{90.30} & 12.43 & 16.69 & \textbf{87.09} & 33.79 & 43.81 & 77.28 & 35.70 & 44.07 & 88.10 & 21.32 & 30.25 & 84.93 & 20.26 & 25.62 & \textbf{85.54} & 24.70 & 32.09 \\
AttacKG+ & 26.35 & 19.54 & 16.67 & 25.65 & 36.27 & 26.55 & 26.50 & 25.66 & 20.06 & 48.22 & 13.64 & 15.04 & 48.47 & 44.01 & 42.95 & 35.04 & 27.82 & 24.25 \\
KnowGL & 25.83 & 0.40 & 0.53 & 27.34 & 1.42 & 2.35 & 23.71 & 1.93 & 2.20 & 28.94 & 2.05 & 2.85 & 28.80 & 5.58 & 6.89 & 26.93 & 2.28 & 2.96 \\
\bottomrule
\end{tabular}
}
\end{table*}

Table~\ref{tab:rq1-main} reports per-source precision, recall, and F1 on the five sources. In the Avg columns, precision and recall are arithmetic means across the five sources, and Avg F1 is the arithmetic mean of the five per-source F1 values. Under this aggregation, Task-bank achieves the best source-averaged recall at 64.91\%. Because recall measures how much ground-truth threat knowledge is recovered, CTINexus's lower average recall (63.47\%) means that it misses more critical cybersecurity information overall. CTINexus attains a nearly identical Avg F1 (68.66\% vs.\ 68.53\%) and slightly higher Avg precision (85.52\% vs.\ 84.62\%), but its strongest result appears on the CTINexus source itself, which is also the shortest source in our suite at only 191 tokens on average. By contrast, if we exclude the CTINexus source itself and re-average only over the other four sources, Task-bank has higher average recall and Avg F1. The efficiency gap also matters: in the standard design of CTINexus, the method uses five inference steps, whereas GRID uses a lower-cost fixed two-prompt pipeline. As a result, GRID's inference token cost is roughly 40\% of the similarly strong CTINexus pipeline~\cite{cheng2025ctinexus}. End2End Reward reaches 58.06\% Avg F1. GraphRAG still attains the highest Avg precision (85.54\%), but its Avg F1 drops to 32.09\% because recall collapses. Across the five sources, both GRID variants remain substantially stronger than almost all other baselines. In particular, Task-bank achieves the best recall and F1 on both the GRID source (78.11\%, 79.62\%) and CASIE (77.68\%, 77.22\%), and the best precision on SecureNLP (85.97\%).

\subsection{RQ2: Post-Training Variants in the GRID Framework}
\begin{table*}[t]
\centering
\footnotesize
\setlength{\tabcolsep}{4pt}
\caption{Post-training variants in GRID across five sources}
\label{tab:rq2-main}
\begin{tabular}{lccc}
\toprule
\textbf{Post-training design} & \textbf{Precision (\%)} & \textbf{Recall (\%)} & \textbf{F1 (\%)} \\
\midrule
Task-bank & \textbf{84.62} & \textbf{64.91} & \textbf{68.53} \\
End2End & 76.91 & 53.85 & 58.06 \\
Choice-only & 72.03 & 33.73 & 40.56 \\
End2End SFT without RL & 69.75 & 30.48 & 37.13 \\
No post-training & 73.99 & 27.21 & 35.44 \\
\bottomrule
\end{tabular}
\end{table*}

Table~\ref{tab:rq2-main} compares five post-training variants in the GRID framework. Here, precision and recall denote five-source averages, and F1 denotes the arithmetic mean of the five per-source F1 values. The first two rows are the two main \tool systems already shown in RQ1: the primary Task-bank Reward model and the secondary End2End Reward model. The remaining three rows are weaker variants built around them: Choice-only Reward without the regex branch, End2End SFT without RL, and the original No post-training model. These ablations further validate Task-bank Reward as an effective reward design. It remains the strongest setting within GRID at 68.53\% Avg F1 and 64.91\% source-averaged recall, outperforming End2End Reward at 58.06\% Avg F1 as well as Choice-only Reward, End2End SFT without RL, and No post-training at 40.56\%, 37.13\%, and 35.44\%, respectively. In particular, compared with Choice-only Reward, Task-bank Reward improves source-averaged recall by 31.18 recall points, showing that the regex branch effectively boosts recall.

Although End2End Reward remains strong, it still trails Task-bank Reward by 11.06 recall points and 10.47 F1 points while incurring much higher online judge cost. Task-bank's one-time offline cost is about \$60 in total, including about \$33 for multi-select question generation and \$27 for regex-target generation; the resulting supervision bank can then be reused across later SFT and RL runs. End2End Reward incurs about \$942 of online LLM-as-judge cost, and this cost grows further as the number of RL steps increases. The concrete End2End training-cost calculation is summarized in Appendix~\ref{app:training-details}.

\subsection{RQ3: Ablation of Article Rewriting and Article-Complexity-Ordered Training}

RQ3 studies two design choices in the Task-bank Reward pipeline: article rewriting and article-complexity-ordered training. We compare three variants under matched RL settings with the same original Qwen3-4B-Instruct-2507 model, 500 training articles, supervision tasks, reward function, and optimization hyperparameters. Evaluation uses a fixed 25-article subset (five per source; seed 42). The variants differ only in whether supervision is built from revised or raw article--KG pairs and whether article blocks follow article-complexity order or random order. We report the training reward together with the small-evaluation-set score, defined as the average of precision and recall:
\begin{itemize}[noitemsep, topsep=2pt, partopsep=1pt, listparindent=\parindent, leftmargin=*]
    \item \textbf{Full setting}: revised-article supervision with article-complexity ordering.
    \item \textbf{w/o article-complexity ordering}: the same supervision, but random article-block order.
    \item \textbf{w/o article rewriting}: supervision regenerated from raw article--KG pairs, with article-complexity ordering retained.
\end{itemize}

\begin{table}[t]
\centering
\footnotesize
\setlength{\tabcolsep}{5pt}
\caption{RQ3 ablation under a shared 225-step training budget}
\label{tab:rq3-main}
\begin{tabular}{lcc}
\toprule
\textbf{Setting} & \textbf{Train reward} & \textbf{Test score $(P+R)/2$} \\
\midrule
Full setting & \textbf{0.7917} & \textbf{0.6641} \\
w/o article rewriting & 0.6265 & 0.6371 \\
w/o article-complexity ordering & 0.3906 & 0.5025 \\
\bottomrule
\end{tabular}
\end{table}

\begin{figure}[t]
\centering
\includegraphics[width=\columnwidth]{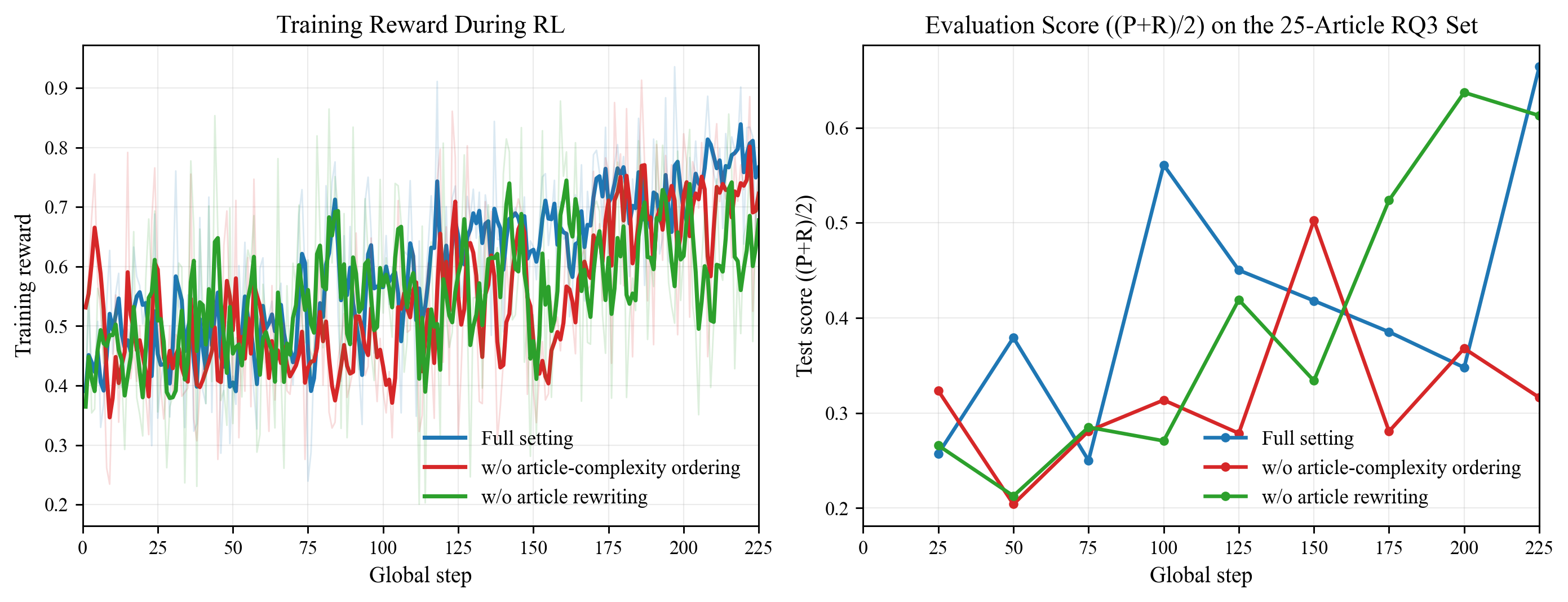}
\caption{RQ3 ablation curves up to 225 steps. Left: RL training reward (first-order EMA, smoothing weight 0.6). Right: test score $(P+R)/2$ on the fixed 25-article RQ3 set}
\label{fig:rq3-curves}
\end{figure}

At the shared 225-step budget, the full setting reaches the best test score $(P+R)/2$ of 0.6641, compared with 0.6371 for w/o article rewriting and 0.5025 for w/o article-complexity ordering. This shows that, without manual supervision, both second-pass article revision and article-complexity-ordered training improve RL reward and test-set precision and recall.

\section{Discussion}
\label{sec:discussion}

We did not train an agent-loop extractor because agent-loop RL requires a more complex reward interface than \tool and current open-source stacks remain immature for stable optimization. We therefore focus on a non-agent extractor in this work.

We chose Qwen3-4B-Instruct-2507~\cite{qwen4b2507hf} because 8B and 14B full-parameter post-training is much slower on our 4$\times$RTX 6000 Ada setup and cannot sustain prompt and response lengths as long as those used for the 4B model, and a recent fine-tuning report ranks it ahead of Llama-3.1-8B-Instruct and Llama-3.2-3B-Instruct. Notably, the latest publicly released small Llama checkpoint still dates back to September 2024~\cite{distillabs2025slmft,llama31hf,llama32hf}.

\section{Related Work}
\label{sec:literature}


\noindent\textbf{Security knowledge graph extraction.}
Methods range from supervised pipelines such as EXTRACTOR~\cite{satvat2021extractor} to LLM-based systems such as CTIKG~\cite{huang2024ctikg}, CTINexus~\cite{cheng2025ctinexus}, CTI-Thinker~\cite{yang2026cti}, LLM-TIKG~\cite{hu2024llm}, LLM-CAKG~\cite{wang2025automated}, and AttacKG+~\cite{zhang2025attackg+}. Most optimize only one stage of the pipeline and rely on general-purpose large commercial LLMs instead of economical small open-source models. In contrast, \tool learns from large volumes of security text to equip the extractor itself with security-domain knowledge.

\noindent\textbf{RL-based relation extraction.}
R1-RE~\cite{dai2025r1} introduces RLVR into relation extraction, but its main task is relation classification over given entity pairs rather than open triplet extraction or complete article-level knowledge graph extraction from full security reports.

\section{Conclusion}

We presented \tool, a CTI knowledge graph extraction framework combining article--graph alignment, task-bank post-training, and ontology-guided inference. On a unified CTI test set spanning five sources, the Task-bank Reward model with GRID inference achieved the best source-averaged recall and a near-tied top Avg F1 at about 40\% of the inference tokens of the similarly strong CTINexus pipeline; RQ2/RQ3 further showed reusable task-bank supervision and gains from KG-conditioned article rewriting plus article-complexity ordering.

\bibliography{./bibliography/refs}

\begin{thebibliography}{70}
\providecommand{\natexlab}[1]{#1}
\providecommand{\url}[1]{\texttt{#1}}
\expandafter\ifx\csname urlstyle\endcsname\relax
  \providecommand{\doi}[1]{doi: #1}\else
  \providecommand{\doi}{doi: \begingroup \urlstyle{rm}\Url}\fi

\bibitem[Vir()]{VirusTotal}
Virustotal - free online virus, malware and url scanner.
\newblock \url{https://www.virustotal.com/en/}.

\bibitem[cyb(2021)]{cyberkillchain}
Cyber kill chain, 2021.
\newblock
  https://www.lockheedmartin.com/en-us/capabilities/cyber/cyber-kill-chain.html.

\bibitem[dep(2021)]{depimpact}
{DepImpact Project Website}, 2021.
\newblock https://github.com/usenixsub/DepImpact.

\bibitem[gri(2026)]{gridrepo}
Projectgrid repository, 2026.
\newblock https://github.com/anonymousauthorname/ProjectGRID.

\bibitem[Bi et~al.(2024)Bi, Chen, Jiang, Xiong, Guo, Chen, and
  Zhang]{bi2024codekgc}
Zhen Bi, Jing Chen, Yinuo Jiang, Feiyu Xiong, Wei Guo, Huajun Chen, and Ningyu
  Zhang.
\newblock Codekgc: Code language model for generative knowledge graph
  construction.
\newblock \emph{ACM Transactions on Asian and Low-Resource Language Information
  Processing}, 23\penalty0 (3):\penalty0 1--16, 2024.

\bibitem[Catakoglu et~al.(2016)Catakoglu, Balduzzi, and Balzarotti]{ioc3}
Onur Catakoglu, Marco Balduzzi, and Davide Balzarotti.
\newblock Automatic extraction of indicators of compromise for web
  applications.
\newblock In \emph{Proceedings of the 25th international conference on world
  wide web}, pp.\  333--343, 2016.

\bibitem[Cheng et~al.(2025)Cheng, Bajaber, Tsegai, Song, and
  Gao]{cheng2025ctinexus}
Yutong Cheng, Osama Bajaber, Saimon~Amanuel Tsegai, Dawn Song, and Peng Gao.
\newblock Ctinexus: Automatic cyber threat intelligence knowledge graph
  construction using large language models.
\newblock In \emph{2025 IEEE 10th European Symposium on Security and Privacy
  (EuroS\&P)}, pp.\  923--938. IEEE, 2025.

\bibitem[Chhikara et~al.(2025)Chhikara, Khant, Aryan, Singh, and
  Yadav]{chhikara2025mem0}
Prateek Chhikara, Dev Khant, Saket Aryan, Taranjeet Singh, and Deshraj Yadav.
\newblock Mem0: Building production-ready ai agents with scalable long-term
  memory.
\newblock \emph{arXiv preprint arXiv:2504.19413}, 2025.

\bibitem[{Cisco}(2024)]{ciscocyberattacks}
{Cisco}.
\newblock Common types of cyber attacks, 2024.
\newblock URL
  \url{https://www.cisco.com/c/en/us/products/security/common-cyberattacks.html}.

\bibitem[{Cognee Contributors}(2025)]{cogneerepo}
{Cognee Contributors}.
\newblock Cognee, 2025.
\newblock URL \url{https://github.com/topoteretes/cognee}.
\newblock GitHub repository.

\bibitem[{Conference on Language Modeling}(2026)]{colm2026cfp}
{Conference on Language Modeling}.
\newblock {COLM} 2026: Call for papers.
\newblock \url{https://colmweb.org/cfp.html}, 2026.
\newblock Accessed: 2026-03-29.

\bibitem[Corporation(2022)]{mitreattack}
The~MITRE Corporation.
\newblock Mitre att\&ck, 2022.
\newblock https://attack.mitre.org/.

\bibitem[{CrowdStrike}(2024)]{crowdstrike2024threatreport}
{CrowdStrike}.
\newblock Crowdstrike 2024 global threat report, 2024.
\newblock URL \url{https://www.crowdstrike.com/global-threat-report/}.

\bibitem[Dai et~al.(2025)Dai, Zheng, Yang, and Zhu]{dai2025r1}
Runpeng Dai, Tong Zheng, Run Yang, and Hongtu Zhu.
\newblock R1-re: Cross-domain relationship extraction with rlvr.
\newblock \emph{arXiv e-prints}, pp.\  arXiv--2507, 2025.

\bibitem[{distil labs}(2025)]{distillabs2025slmft}
{distil labs}.
\newblock We benchmarked 12 small language models across 8 tasks to find the
  best base model for fine-tuning.
\newblock \url{https://www.distillabs.ai/blog/}, December 2025.
\newblock Published: 2025-12-10; accessed: 2026-03-29.

\bibitem[Dong et~al.(2019)Dong, Guo, Chen, Xing, Zhang, and
  Wang]{dong2019towards}
Ying Dong, Wenbo Guo, Yueqi Chen, Xinyu Xing, Yuqing Zhang, and Gang Wang.
\newblock Towards the detection of inconsistencies in public security
  vulnerability reports.
\newblock In \emph{28th USENIX security symposium (USENIX Security 19)}, pp.\
  869--885, 2019.

\bibitem[Gao et~al.(2025)Gao, Zheng, Chen, Dang, Liu, Yu, Yang, Bai, Zhou, and
  Lin]{gao2025soft}
Chang Gao, Chujie Zheng, Xiong-Hui Chen, Kai Dang, Shixuan Liu, Bowen Yu,
  An~Yang, Shuai Bai, Jingren Zhou, and Junyang Lin.
\newblock Soft adaptive policy optimization.
\newblock \emph{arXiv preprint arXiv:2511.20347}, 2025.

\bibitem[Gao et~al.(2018)Gao, Xiao, Li, Xu, Kulkarni, and Mittal]{gao2018aiql}
Peng Gao, Xusheng Xiao, Zhichun Li, Fengyuan Xu, Sanjeev~R. Kulkarni, and
  Prateek Mittal.
\newblock {AIQL}: Enabling efficient attack investigation from system
  monitoring data.
\newblock In \emph{USENIX Annual Technical Conference (ATC)}, pp.\  113--126,
  2018.

\bibitem[Gao et~al.(2022)Gao, Liu, Choi, Ma, Yang, Ji, Zhang, and
  Song]{gao2022threatkg}
Peng Gao, Xiaoyuan Liu, Edward Choi, Sibo Ma, Xinyu Yang, Zhengjie Ji, Zilin
  Zhang, and Dawn Song.
\newblock Threatkg: A threat knowledge graph for automated open-source cyber
  threat intelligence gathering and management.
\newblock \emph{arXiv preprint arXiv:2212.10388}, 2022.

\bibitem[He et~al.(2024)He, Tian, Sun, Chawla, Laurent, LeCun, Bresson, and
  Hooi]{he2024g}
Xiaoxin He, Yijun Tian, Yifei Sun, Nitesh~V Chawla, Thomas Laurent, Yann LeCun,
  Xavier Bresson, and Bryan Hooi.
\newblock G-retriever: Retrieval-augmented generation for textual graph
  understanding and question answering.
\newblock \emph{Advances in Neural Information Processing Systems},
  37:\penalty0 132876--132907, 2024.

\bibitem[Hu et~al.(2024)Hu, Zou, Han, Sun, and Wang]{hu2024llm}
Yuelin Hu, Futai Zou, Jiajia Han, Xin Sun, and Yilei Wang.
\newblock Llm-tikg: Threat intelligence knowledge graph construction utilizing
  large language model.
\newblock \emph{Computers \& Security}, 145:\penalty0 103999, 2024.

\bibitem[Huang \& Xiao(2024)Huang and Xiao]{huang2024ctikg}
Liangyi Huang and Xusheng Xiao.
\newblock Ctikg: Llm-powered knowledge graph construction from cyber threat
  intelligence.
\newblock In \emph{First Conference on Language Modeling}, 2024.

\bibitem[Huguet~Cabot \& Navigli(2021)Huguet~Cabot and
  Navigli]{huguet-cabot-navigli-2021-rebel-relation}
Pere-Llu{\'\i}s Huguet~Cabot and Roberto Navigli.
\newblock {REBEL}: Relation extraction by end-to-end language generation.
\newblock In \emph{Findings of the Association for Computational Linguistics:
  EMNLP 2021}, pp.\  2370--2381, Punta Cana, Dominican Republic, November 2021.
  Association for Computational Linguistics.
\newblock URL \url{https://aclanthology.org/2021.findings-emnlp.204}.

\bibitem[Hutchins et~al.(2011)Hutchins, Cloppert, and Amin]{apt}
Eric~M Hutchins, Michael~J Cloppert, and Rohan~M Amin.
\newblock Intelligence-driven computer network defense informed by analysis of
  adversary campaigns and intrusion kill chains.
\newblock \emph{Leading Issues in Information Warfare \& Security Research},
  1:\penalty0 80, 2011.

\bibitem[King \& Chen(2003)King and Chen]{backtracking}
Samuel~T. King and Peter~M. Chen.
\newblock Backtracking intrusions.
\newblock In \emph{ACM Symposium on Operating systems principles (SOSP)}, pp.\
  223--236. ACM, 2003.

\bibitem[Li et~al.(2024)Li, Shi, Pan, Zhao, and Sun]{li2024cybersecurity}
Hongyi Li, Ze~Shi, Chengwei Pan, Di~Zhao, and Nan Sun.
\newblock Cybersecurity knowledge graphs construction and quality assessment.
\newblock \emph{Complex \& Intelligent Systems}, 10\penalty0 (1):\penalty0
  1201--1217, 2024.

\bibitem[Li et~al.(2022)Li, Zeng, Chen, and Liang]{li2022attackg}
Zhenyuan Li, Jun Zeng, Yan Chen, and Zhenkai Liang.
\newblock Attackg: Constructing technique knowledge graph from cyber threat
  intelligence reports.
\newblock In \emph{European Symposium on Research in Computer Security}, pp.\
  589--609. Springer, 2022.

\bibitem[Liao et~al.(2016)Liao, Yuan, Wang, Li, Xing, and Beyah]{ioc2}
Xiaojing Liao, Kan Yuan, XiaoFeng Wang, Zhou Li, Luyi Xing, and Raheem Beyah.
\newblock Acing the ioc game: Toward automatic discovery and analysis of
  open-source cyber threat intelligence.
\newblock In \emph{Proceedings of the 2016 ACM SIGSAC conference on computer
  and communications security}, pp.\  755--766, 2016.

\bibitem[Lim et~al.(2017)Lim, Muis, Lu, and Ong]{lim2017malwaretextdb}
Swee~Kiat Lim, Aldrian~Obaja Muis, Wei Lu, and Chen~Hui Ong.
\newblock Malwaretextdb: A database for annotated malware articles.
\newblock In \emph{Proceedings of the 55th Annual Meeting of the Association
  for Computational Linguistics (Volume 1: Long Papers)}, pp.\  1557--1567.
  Association for Computational Linguistics, 2017.
\newblock \doi{10.18653/v1/P17-1143}.
\newblock URL \url{https://aclanthology.org/P17-1143/}.

\bibitem[Liu et~al.(2024)Liu, Wang, Ding, Li, and Zhang]{liu2024bvted}
Kai Liu, Yi~Wang, Zhaoyun Ding, Aiping Li, and Weiming Zhang.
\newblock Bvted: A specialized bilingual (chinese--english) dataset for
  vulnerability triple extraction tasks.
\newblock \emph{Applied Sciences}, 14\penalty0 (16):\penalty0 7310, 2024.

\bibitem[{Mandiant}(2024)]{mandiant2024global}
{Mandiant}.
\newblock Global perspectives on threat intelligence, 2024.
\newblock URL
  \url{https://assets.starlinkme.net/gitex-vendor-assets/mandiant/Global%20Perspectives%20on%20Threat%20Intelligence.pdf}.

\bibitem[McMillan(2013)]{cti}
Rob McMillan.
\newblock Open threat intelligence, 2013.
\newblock https://www.gartner.com/doc/2487216/definition-threat-intelligence.

\bibitem[{Meta}(2024{\natexlab{a}})]{llama31hf}
{Meta}.
\newblock Llama-3.1-8b-instruct.
\newblock \url{https://huggingface.co/meta-llama/Llama-3.1-8B-Instruct}, July
  2024{\natexlab{a}}.
\newblock Released: 2024-07-23; accessed: 2026-03-29.

\bibitem[{Meta}(2024{\natexlab{b}})]{llama32hf}
{Meta}.
\newblock Llama-3.2-3b.
\newblock \url{https://huggingface.co/meta-llama/Llama-3.2-3B}, September
  2024{\natexlab{b}}.
\newblock Released: 2024-09-25, accessed: 2026-03-29.

\bibitem[Micro(2023)]{magniber2023}
Trend Micro.
\newblock Ransomware spotlight: Magniber, 2023.
\newblock URL
  \url{https://www.trendmicro.com/vinfo/us/security/news/ransomware-spotlight/ransomware-spotlight-magniber}.

\bibitem[{Microsoft}(2024{\natexlab{a}})]{graphragrepo}
{Microsoft}.
\newblock Graphrag, 2024{\natexlab{a}}.
\newblock URL \url{https://github.com/microsoft/graphrag}.
\newblock GitHub repository.

\bibitem[{Microsoft}(2024{\natexlab{b}})]{microsoftcyberattack}
{Microsoft}.
\newblock What is a cyberattack?, 2024{\natexlab{b}}.
\newblock URL
  \url{https://www.microsoft.com/en-us/security/business/security-101/what-is-a-cyberattack}.

\bibitem[Milajerdi et~al.(2019)Milajerdi, Eshete, Gjomemo, and
  Venkatakrishnan]{poirot}
Sadegh~M. Milajerdi, Birhanu Eshete, Rigel Gjomemo, and V.N. Venkatakrishnan.
\newblock Poirot: Aligning attack behavior with kernel audit records for cyber
  threat hunting.
\newblock In \emph{ACM Conference on Computer and Communications Security
  (CCS)}, pp.\  1795–1812, 2019.

\bibitem[MITRE(2020)]{cve}
MITRE.
\newblock {Common Vulnerabilities and Exposures (CVE)}, 2020.
\newblock https://cve.mitre.org/.

\bibitem[Obrst et~al.(2012)Obrst, Chase, and Markeloff]{ioc}
Leo Obrst, Penny Chase, and Richard Markeloff.
\newblock Developing an ontology of the cyber security domain.
\newblock In \emph{STIDS}, pp.\  49--56. Citeseer, 2012.

\bibitem[of~Standards \& Technology(2021)of~Standards and Technology]{nvd}
National~Institute of~Standards and Technology.
\newblock National vulnerability database (nvd), 2021.
\newblock https://nvd.nist.gov/.

\bibitem[{Office of the National Cyber
  Director}(2024)]{cybersecurityposture2024}
{Office of the National Cyber Director}.
\newblock 2024 report on the cybersecurity posture of the united states, 2024.
\newblock URL
  \url{https://www.whitehouse.gov/wp-content/uploads/2024/05/2024-Report-on-the-Cybersecurity-Posture-of-the-United-States.pdf}.

\bibitem[OpenAI(2023)]{chatgpt}
OpenAI.
\newblock Chatgpt: Applications, opportunities, and threats.
\newblock \emph{arXiv preprint arXiv:2304.09103}, 2023.

\bibitem[{OpenAI}(2026{\natexlab{a}})]{openai2026gpt53codex}
{OpenAI}.
\newblock Introducing gpt-5.3-codex.
\newblock \url{https://openai.com/index/introducing-gpt-5-3-codex/}, February
  2026{\natexlab{a}}.
\newblock February 5, 2026.

\bibitem[{OpenAI}(2026{\natexlab{b}})]{openai2026gpt54mini}
{OpenAI}.
\newblock Introducing gpt-5.4 mini and nano.
\newblock \url{https://openai.com/index/introducing-gpt-5-4-mini-and-nano/},
  March 2026{\natexlab{b}}.
\newblock March 17, 2026.

\bibitem[Phandi et~al.(2018)Phandi, Silva, and Lu]{phandi2018securenlp}
Peter Phandi, Amila Silva, and Wei Lu.
\newblock Semeval-2018 task 8: Semantic extraction from cybersecurity reports
  using natural language processing ({SecureNLP}).
\newblock In \emph{Proceedings of the 12th International Workshop on Semantic
  Evaluation}, pp.\  697--706. Association for Computational Linguistics, 2018.
\newblock \doi{10.18653/v1/S18-1113}.
\newblock URL \url{https://aclanthology.org/S18-1113/}.

\bibitem[{Qwen Team}(2025{\natexlab{a}})]{qwen3tech}
{Qwen Team}.
\newblock Qwen3 technical report.
\newblock \emph{arXiv preprint arXiv:2505.09388}, 2025{\natexlab{a}}.
\newblock \doi{10.48550/arXiv.2505.09388}.
\newblock URL \url{https://arxiv.org/abs/2505.09388}.

\bibitem[{Qwen Team}(2025{\natexlab{b}})]{qwen4b2507hf}
{Qwen Team}.
\newblock Qwen3-4b-instruct-2507.
\newblock \url{https://huggingface.co/Qwen/Qwen3-4B-Instruct-2507},
  2025{\natexlab{b}}.
\newblock Accessed: 2026-03-29.

\bibitem[Rasmussen et~al.()Rasmussen, Paliychuk, Beauvais, Ryan, and
  Chalef]{rasmussen2501zep}
Preston Rasmussen, Pavlo Paliychuk, Travis Beauvais, Jack Ryan, and Daniel
  Chalef.
\newblock Zep: A temporal knowledge graph architecture for agent memory, 2025.
\newblock \emph{URL https://arxiv. org/abs/2501.13956}.

\bibitem[Rastogi et~al.(2021)Rastogi, Dutta, Christian, Zaki, Gittens, and
  Aggarwal]{rastogi2021information}
Nidhi Rastogi, Sharmishtha Dutta, Ryan Christian, Mohammad Zaki, Alex Gittens,
  and Charu Aggarwal.
\newblock Information prediction using knowledge graphs for contextual malware
  threat intelligence.
\newblock \emph{arXiv preprint arXiv:2102.05571}, 2021.
\newblock URL \url{https://arxiv.org/abs/2102.05571}.

\bibitem[Rossiello et~al.(2023)Rossiello, Chowdhury, Mihindukulasooriya,
  Cornec, and Gliozzo]{rossiello2023knowgl}
Gaetano Rossiello, Md~Faisal~Mahbub Chowdhury, Nandana Mihindukulasooriya, Owen
  Cornec, and Alfio~Massimiliano Gliozzo.
\newblock Knowgl: Knowledge generation and linking from text.
\newblock In \emph{Proceedings of the AAAI Conference on Artificial
  Intelligence}, volume~37, pp.\  16476--16478, 2023.

\bibitem[Satvat et~al.(2021)Satvat, Gjomemo, and
  Venkatakrishnan]{satvat2021extractor}
Kiavash Satvat, Rigel Gjomemo, and VN~Venkatakrishnan.
\newblock Extractor: Extracting attack behavior from threat reports.
\newblock In \emph{2021 IEEE European Symposium on Security and Privacy
  (EuroS\&P)}, pp.\  598--615. IEEE, 2021.

\bibitem[Satyapanich et~al.(2020)Satyapanich, Ferraro, and
  Finin]{satyapanich2020casie}
Taneeya Satyapanich, Francis Ferraro, and Tim Finin.
\newblock Casie: Extracting cybersecurity event information from text.
\newblock In \emph{Proceedings of the AAAI conference on artificial
  intelligence}, volume~34, pp.\  8749--8757, 2020.

\bibitem[{Securelist}(2024)]{securelistthreatcategories}
{Securelist}.
\newblock Threat categories, 2024.
\newblock URL \url{https://securelist.com/threat-categories/}.

\bibitem[Senki()]{ctifeed2}
Senki.
\newblock Open source threat intelligence feeds.
\newblock
  https://www.senki.org/operators-security-toolkit/open-source-threat-intelligence-feeds/.

\bibitem[Senki(2016)]{ctifeed1}
Senki.
\newblock Real-time threat intelligence, 2016.
\newblock https://www.recordedfuture.com/.

\bibitem[Sheng et~al.(2024)Sheng, Zhang, Ye, Wu, Zhang, Zhang, Peng, Lin, and
  Wu]{sheng2024hybridflow}
Guangming Sheng, Chi Zhang, Zilingfeng Ye, Xibin Wu, Wang Zhang, Ru~Zhang,
  Yanghua Peng, Haibin Lin, and Chuan Wu.
\newblock Hybridflow: A flexible and efficient rlhf framework.
\newblock \emph{arXiv preprint arXiv: 2409.19256}, 2024.

\bibitem[Sun et~al.(2023{\natexlab{a}})Sun, Xu, Tang, Wang, Lin, Gong, Ni,
  Shum, and Guo]{sun2023think}
Jiashuo Sun, Chengjin Xu, Lumingyuan Tang, Saizhuo Wang, Chen Lin, Yeyun Gong,
  Lionel~M Ni, Heung-Yeung Shum, and Jian Guo.
\newblock Think-on-graph: Deep and responsible reasoning of large language
  model on knowledge graph, 2024.
\newblock \emph{URL https://arxiv. org/abs/2307.07697}, 2023{\natexlab{a}}.

\bibitem[Sun et~al.(2023{\natexlab{b}})Sun, Huang, Wu, and Shen]{sun2023sectkg}
Siqi Sun, Cheng Huang, Tiejun Wu, and Yi~Shen.
\newblock Sectkg: A knowledge graph for open-source security tools.
\newblock \emph{International Journal of Intelligent Systems}, 2023\penalty0
  (1):\penalty0 4464974, 2023{\natexlab{b}}.

\bibitem[Times(2014)]{target}
New~York Times.
\newblock {Target data breach incident}, 2014.
\newblock
  http://www.nytimes.com/2014/02/27/business/target-reports-on-fourth-quarter-earnings.html?\_r=1.

\bibitem[{Unit42 by Palo Alto Networks}(2017)]{eltestcampaigns}
{Unit42 by Palo Alto Networks}.
\newblock How the eitest campaigns path to angler ek evolved over time, 2017.
\newblock URL
  \url{https://unit42.paloaltonetworks.com/unit42-how-the-eltest-campaigns-path-to-angler-ek-evolved-over-time/}.

\bibitem[Wagner et~al.(2019)Wagner, Mahbub, Palomar, and Abdallah]{cti2}
Thomas~D Wagner, Khaled Mahbub, Esther Palomar, and Ali~E Abdallah.
\newblock Cyber threat intelligence sharing: Survey and research directions.
\newblock \emph{Computers \& Security}, 87:\penalty0 101589, 2019.

\bibitem[Wang et~al.(2023)Wang, Zhang, Li, and Liu]{llms}
Y.~Wang, Y.~Zhang, Y.~Li, and X.~Liu.
\newblock A bibliometric review of large language models research from 2017 to
  2023.
\newblock \emph{arXiv preprint arXiv:2304.02020}, 2023.

\bibitem[Wang et~al.(2025)Wang, Fei, Hu, Shan, Xiao, You, and
  Chen]{wang2025automated}
Zhihua Wang, Siyuan Fei, Youlin Hu, Dacheng Shan, Shitao Xiao, Lizhao You, and
  Peijun Chen.
\newblock Automated attack knowledge graph construction with large language
  models.
\newblock In \emph{Proceedings of the 2025 2nd International Conference on
  Computer and Multimedia Technology}, pp.\  700--706, 2025.

\bibitem[Xu et~al.(2022)Xu, Fang, Liu, Xiao, Wen, and Meng]{depcomm}
Zhiqiang Xu, Pengcheng Fang, Changlin Liu, Xusheng Xiao, Yu~Wen, and Dan Meng.
\newblock Depcomm: Graph summarization on system audit logs for attack
  investigation.
\newblock In \emph{2022 IEEE Symposium on Security and Privacy (SP)}, pp.\
  540--557. IEEE, 2022.

\bibitem[Yang et~al.(2026)Yang, Zhong, Chen, Peng, Yao, Chen, Wang, Zhang,
  Zhou, and Yang]{yang2026cti}
Xiuzhang Yang, Ruijie Zhong, Yuling Chen, Guojun Peng, Di~Yao, Chaofan Chen,
  Chenyang Wang, Dongni Zhang, Yilin Zhou, and Zixuan Yang.
\newblock Cti-thinker: an llm-driven system for cti knowledge graph
  construction and attack reasoning.
\newblock \emph{Cybersecurity}, 9\penalty0 (1):\penalty0 106, 2026.

\bibitem[Yu et~al.(2025)Yu, Chiang, Tsai, Huang, and Tsao]{primus}
Yao-Ching Yu, Tsun-Han Chiang, Cheng-Wei Tsai, Chien-Ming Huang, and Wen-Kwang
  Tsao.
\newblock Primus: A pioneering collection of open-source datasets for
  cybersecurity llm training.
\newblock \emph{arXiv preprint arXiv:2502.11191}, 2025.
\newblock URL \url{https://arxiv.org/abs/2502.11191}.

\bibitem[Yue et~al.(2024)Yue, Wang, Chen, Jiang, Fu, and Jiang]{yue2024hrtc}
HuanZhou Yue, XuRen Wang, Rong Chen, ZhengWei Jiang, YuXia Fu, and Jun Jiang.
\newblock Hrtc: A triplet joint extraction model based on cyber threat
  intelligence.
\newblock In \emph{International Conference on Knowledge Science, Engineering
  and Management}, pp.\  214--223. Springer, 2024.

\bibitem[{Zep}(2025)]{graphitirepo}
{Zep}.
\newblock Graphiti, 2025.
\newblock URL \url{https://github.com/getzep/graphiti}.
\newblock GitHub repository.

\bibitem[Zhang et~al.(2025)Zhang, Du, Ma, Wang, Xie, Yang, Lu, and
  Chang]{zhang2025attackg+}
Yongheng Zhang, Tingwen Du, Yunshan Ma, Xiang Wang, Yi~Xie, Guozheng Yang,
  Yuliang Lu, and Ee-Chien Chang.
\newblock Attackg+: Boosting attack graph construction with large language
  models.
\newblock \emph{Computers \& Security}, 150:\penalty0 104220, 2025.

\end{thebibliography}
\bibliographystyle{colm2026conference}
\clearpage
\appendix
\section{LLM Usage Disclosure}
\label{app:llm-usage}

In line with the COLM 2026 policy on LLM use disclosure~\cite{colm2026cfp}, we disclose the non-minor LLM usage in this work's research pipeline. In training-data construction, we use GPT-5.3 Codex Medium~\cite{openai2026gpt53codex} to generate traceable article-to-KG alignments, perform KG-conditioned text revision, create four-option multi-select questions, generate triple-level regex targets, and compute the training-side precision/recall reward used by the End2End setting. In evaluation on the human-annotated benchmark articles, we use GPT-5.4 mini~\cite{openai2026gpt54mini} as an LLM-as-judge to measure the effectiveness of different methods by computing precision and recall for their generated knowledge graphs against human-annotated ground-truth graphs.

\section{Security Ontology Inventory}
\label{app:ontology}

The ontology used by \tool defines both the structured fields attached to graph elements and the controlled vocabularies used during extraction and validation. Each entity records \texttt{name}, \texttt{type}, \texttt{alias}, and \texttt{parent entity}. Each relation records \texttt{sub}, \texttt{rel}, \texttt{rel\_type}, and \texttt{obj}. The predefined inventories below follow the GRID ontology used in this paper.

Our ontology is inspired by STIX rather than being a direct copy. During manual labeling, we gradually found that CTI articles repeatedly required several practical refinements beyond the raw STIX inventory. The following types are introduced:

{\raggedright
Merged attacker/activity labels. \texttt{threat-actor-or-intrusion-set}\\
Offensive tools vs.\ legitimate-but-abused software. \texttt{hacker-tool}, \texttt{general-software}\\
Component-level types. \texttt{detailed-part-of-malware-or-hackertool}\\
\texttt{detailed-part-of-general-software}\\
Analysis- or fallback-oriented types. \texttt{security-product},\\
\texttt{malware-analysis-document-}\\
\texttt{or-publication-or-conference}, \texttt{abstract-concept},\\
\texttt{generic-noun}, \texttt{noise}\par}

These types are introduced to guide the LLM toward deeper structural relations among entities in CTI articles, rather than to turn extraction into a closed label-only task. The extraction remains fundamentally open-ended, with the ontology serving as a lightweight scaffold for reasoning and normalization.

\begin{figure}[t]
\centering
\includegraphics[width=0.95\columnwidth]{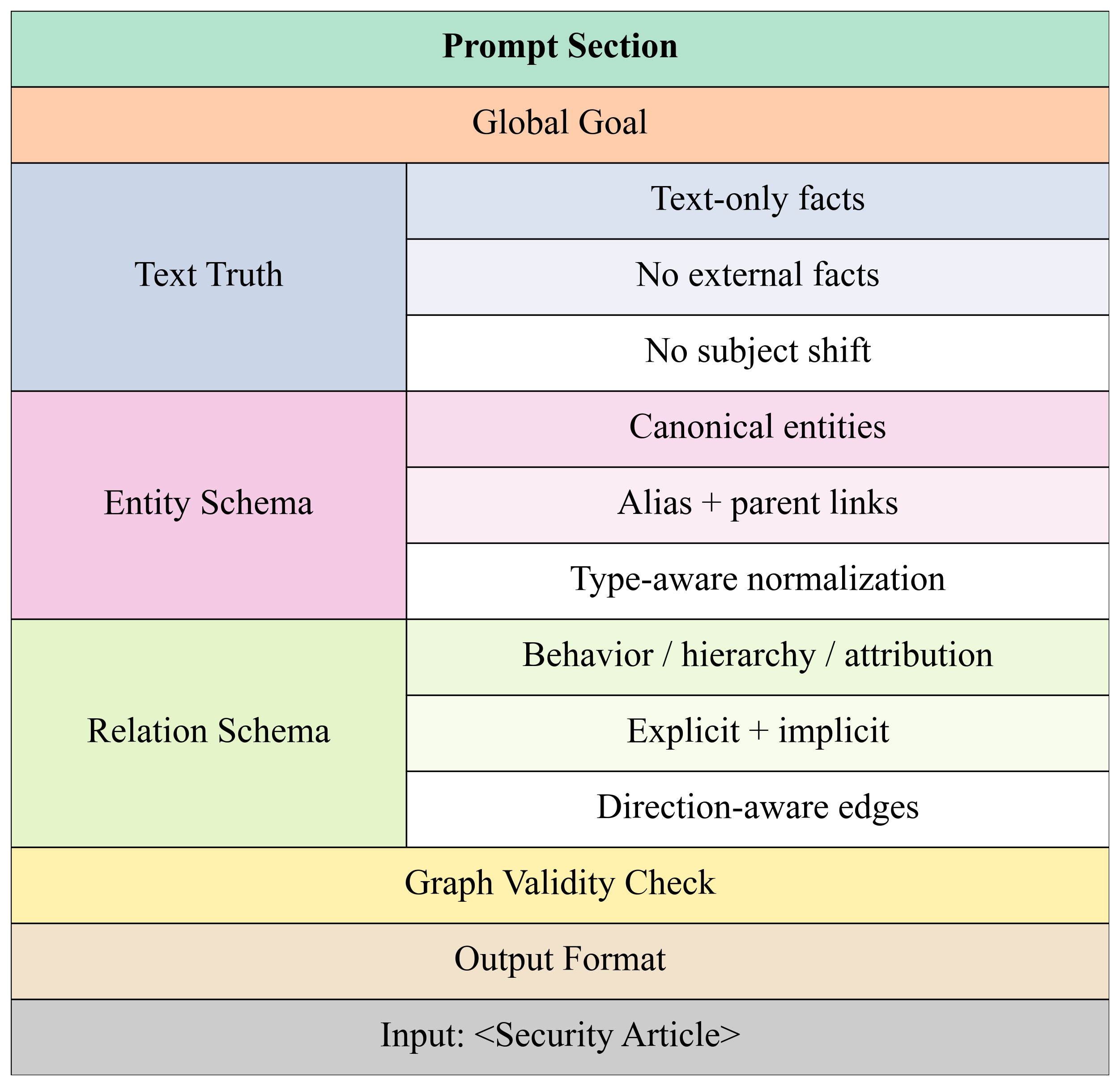}
\caption{Prompt blocks for the \tool extractor}
\label{fig:ontology-prompt-block}
\end{figure}

\begin{table*}[t]
\centering
\footnotesize
\setlength{\tabcolsep}{4pt}
\caption{Entity types in the GRID ontology}
\label{tab:ontology-entities}
\begin{tabular}{p{0.23\textwidth}p{0.23\textwidth}p{0.23\textwidth}p{0.23\textwidth}}
\toprule
\texttt{user-account} & \texttt{identity} & \makecell[l]{\texttt{threat-actor-}\\\texttt{or-intrusion-set}} & \texttt{campaign} \\
\texttt{malware} & \texttt{hacker-tool} & \texttt{general-software} & \texttt{security-product} \\
\makecell[l]{\texttt{detailed-part-of-}\\\texttt{malware-or-}\\\texttt{hackertool}} & \makecell[l]{\texttt{detailed-part-of-}\\\texttt{general-software}} & \texttt{attack-pattern} & \texttt{vulnerability} \\
\texttt{file} & \texttt{process} & \makecell[l]{\texttt{windows-}\\\texttt{registry-key}} & \texttt{course-of-action} \\
\texttt{url} & \texttt{domain-name} & \texttt{ipv4-addr} & \texttt{ipv6-addr} \\
\makecell[l]{\texttt{network-}\\\texttt{traffic}} & \texttt{infrastructure} & \makecell[l]{\texttt{email-}\\\texttt{address}} & \texttt{mac-address} \\
\texttt{indicator} & \makecell[l]{\texttt{malware-analysis-}\\\texttt{document-or-}\\\texttt{publication-}\\\texttt{or-conference}} & \makecell[l]{\texttt{credential-}\\\texttt{value}} & \makecell[l]{\texttt{x509-}\\\texttt{certificate}} \\
\texttt{location} & \makecell[l]{\texttt{abstract-}\\\texttt{concept}} & \texttt{generic-noun} & \texttt{other} \\
\texttt{noise} &  &  &  \\
\bottomrule
\end{tabular}
\end{table*}

\begin{table*}[t]
\centering
\footnotesize
\setlength{\tabcolsep}{4pt}
\caption{Relation types in the GRID ontology}
\label{tab:ontology-relations}
\begin{tabular}{p{0.23\textwidth}p{0.23\textwidth}p{0.23\textwidth}p{0.23\textwidth}}
\toprule
\texttt{exploits} & \texttt{bypasses} & \makecell[l]{\texttt{malicious-investigates-}\\\texttt{track-detects}} & \texttt{impersonates} \\
\texttt{targets} & \texttt{compromises} & \texttt{leads-to} & \texttt{drops} \\
\texttt{downloads} & \texttt{executes} & \texttt{delivers} & \texttt{beacons-to} \\
\texttt{exfiltrate-to} & \texttt{leaks} & \makecell[l]{\texttt{communicates-}\\\texttt{with}} & \texttt{resolves-to} \\
\texttt{hosts} & \texttt{provides} & \texttt{authored-by} & \texttt{owns} \\
\texttt{controls} & \texttt{attributed-to} & \makecell[l]{\texttt{affiliated-}\\\texttt{with}} & \makecell[l]{\texttt{cooperates-}\\\texttt{with}} \\
\texttt{is-part-of} & \texttt{consists-of} & \texttt{has} & \texttt{depends-on} \\
\makecell[l]{\texttt{creates-or-}\\\texttt{generates}} & \makecell[l]{\texttt{modifies-or-}\\\texttt{removes-or-}\\\texttt{replaces}} & \texttt{uses} & \texttt{variant-of} \\
\texttt{derived-from} & \texttt{alias-of} & \texttt{compares-to} & \texttt{categorized-as} \\
\texttt{located-at} & \makecell[l]{\texttt{originates-}\\\texttt{from}} & \texttt{indicates} & \texttt{mitigates} \\
\texttt{based-on} & \makecell[l]{\texttt{research-describes-}\\\texttt{analysis-of-}\\\texttt{characterizes-}\\\texttt{detects}} & \texttt{negation} & \texttt{other} \\
\bottomrule
\end{tabular}
\end{table*}

\section{Judge Rules for Automatic Evaluation}
\label{app:judge-rules}

The automatic evaluator scores precision and recall under an explicit rule set rather than unconstrained semantic similarity. Table~\ref{tab:judge-rules} summarizes the main judge rules.

\begin{table*}[t]
\centering
\footnotesize
\setlength{\tabcolsep}{4pt}
\caption{Judge rules in the automatic evaluator}
\label{tab:judge-rules}
\begin{tabular}{p{0.25\textwidth}p{0.67\textwidth}}
\toprule
\textbf{Rule} & \textbf{Meaning} \\
\midrule
Text-Provable Truth & A relation counts only if it can be supported by the article text itself. The judge must not use external world knowledge, domain defaults, or subject elevation to justify a match. \\
Subject-Preserving Chain Reasoning & Multi-hop reasoning is allowed only when the full chain is text-supported and the subject remains the same throughout the deduction. \\
General-Specific Equivalence & Coarser and finer phrasings may match when they refer to the same fact stated in the text rather than to different entities or events. \\
Action-Technique and Relation Normalization & A surface action, a normalized technique name, and a canonical relation label may match when they denote the same behavior in context. \\
Relation Hierarchy Tolerance & Parent-child relation gaps are tolerated when the article supports the specific behavior and therefore also licenses the coarser relation, or vice versa. \\
Attribute-as-Structure Matching & Structural facts may be represented either as edges or as entity attributes. In particular, \texttt{alias} and \texttt{parent entity} fields can satisfy corresponding ground-truth structural relations. \\
Alias and Hierarchy Equivalence & Alias forms and near-family variants can be treated as equivalent only when the text or entity attributes explicitly support that equivalence. \\
Indexed Edge-wise Auditing & Precision is judged edge by edge over predicted triples, while recall is judged edge by edge over ground-truth triples. For each audited edge, the judge must return its index, a binary decision, and either supporting text evidence or the matched counterpart predicted edge(s). \\
Malformed Extraction Rejection & Predictions whose subject or object is a pronoun, a full clause, or another non-entity span are rejected and cannot count as true positives. \\
\bottomrule
\end{tabular}
\end{table*}

\FloatBarrier
\subsection{Judge Calibration Against Human Annotations}
\label{app:judge-calibration}

To calibrate the indexed edge-wise evaluator, we compare the LLM-as-judge against human judgments on 378 manually reviewed audit items labeled by three human reviewers, including 191 precision items and 187 recall items. The reviewed set spans the methods evaluated in RQ1 and RQ2.

The human interface records whether the annotator agrees or disagrees with the LLM judgment. We therefore convert these decisions into the corresponding human precision labels (TP vs.\ FP) and human recall labels (TP vs.\ FN). For precision, agreeing with an LLM-TP or disagreeing with an LLM-FP implies a human TP, while agreeing with an LLM-FP or disagreeing with an LLM-TP implies a human FP. For recall, the same conversion yields human TP and FN labels.

\begin{center}
\footnotesize
\captionof{table}{Judge--human confusion matrices. Rows are LLM labels and columns are human labels.}
\label{tab:judge-calibration-confusion}
\vspace{4pt}
\begin{minipage}{0.48\columnwidth}
\centering
\textbf{Precision calibration}\\
\vspace{2pt}
\renewcommand{\arraystretch}{1.35}
\setlength{\tabcolsep}{4pt}
\resizebox{0.48\columnwidth}{!}{%
\begin{tabular}{|c|c|c|c|}
\hline
\multicolumn{2}{|c|}{} & \multicolumn{2}{c|}{\makecell{\textbf{Human's}\\\textbf{decision}}} \\
\hline
\multicolumn{2}{|c|}{} & \textbf{TP} & \textbf{FP} \\
\hline
\multirow{2}{*}{\rotatebox{90}{\textbf{LLM}}} & \textbf{TP} & \textbf{87.6\%} & 12.4\% \\
\cline{2-4}
 & \textbf{FP} & 27.9\% & \textbf{72.1\%} \\
\hline
\end{tabular}
}
\end{minipage}
\hfill
\begin{minipage}{0.48\columnwidth}
\centering
\textbf{Recall calibration}\\
\vspace{2pt}
\renewcommand{\arraystretch}{1.35}
\setlength{\tabcolsep}{4pt}
\resizebox{0.48\columnwidth}{!}{%
\begin{tabular}{|c|c|c|c|}
\hline
\multicolumn{2}{|c|}{} & \multicolumn{2}{c|}{\makecell{\textbf{Human's}\\\textbf{decision}}} \\
\hline
\multicolumn{2}{|c|}{} & \textbf{TP} & \textbf{FN} \\
\hline
\multirow{2}{*}{\rotatebox{90}{\textbf{LLM}}} & \textbf{TP} & \textbf{78.6\%} & 21.4\% \\
\cline{2-4}
 & \textbf{FN} & 4.8\% & \textbf{95.2\%} \\
\hline
\end{tabular}
}
\end{minipage}
\end{center}

The resulting agreement is 80.6\% for precision (154/191) and 91.4\% for recall (171/187), yielding an overall agreement of 86.0\% (325/378).

\begin{center}
\scriptsize
\captionof{table}{Per-source agreement rates between the LLM judge and three human reviewers.}
\label{tab:judge-calibration-by-reviewer-source}
\vspace{4pt}
\renewcommand{\arraystretch}{1.2}
\setlength{\tabcolsep}{3pt}
\resizebox{\columnwidth}{!}{%
\begin{tabular}{lcccccc}
\toprule
\textbf{Source} & \multicolumn{2}{c}{\textbf{Reviewer 1}} & \multicolumn{2}{c}{\textbf{Reviewer 2}} & \multicolumn{2}{c}{\textbf{Reviewer 3}} \\
\cmidrule(lr){2-3} \cmidrule(lr){4-5} \cmidrule(lr){6-7}
 & \makecell{\textbf{Precision}\\\textbf{agreement}} & \makecell{\textbf{Recall}\\\textbf{agreement}} & \makecell{\textbf{Precision}\\\textbf{agreement}} & \makecell{\textbf{Recall}\\\textbf{agreement}} & \makecell{\textbf{Precision}\\\textbf{agreement}} & \makecell{\textbf{Recall}\\\textbf{agreement}} \\
\midrule
GRID & 100.0\% & 81.8\% & 77.8\% & 88.9\% & 62.5\% & 100.0\% \\
CASIE & 87.5\% & 80.0\% & 81.8\% & 100.0\% & 100.0\% & 80.0\% \\
CTINexus & 62.5\% & 100.0\% & 80.0\% & 100.0\% & 100.0\% & 100.0\% \\
MalKG & 70.8\% & 100.0\% & 77.3\% & 88.9\% & 84.6\% & 92.9\% \\
SecureNLP & 84.2\% & 93.5\% & 85.7\% & 88.0\% & 77.8\% & 86.7\% \\
\bottomrule
\end{tabular}
}
\end{center}
\FloatBarrier

\section{Baseline Adaptation Details}
\label{app:baseline-details}

For these LLM-based baselines, we first locate and reproduce their public implementations from the corresponding papers, project pages, and code repositories, then apply only the engineering adaptations needed to run them under the same original Qwen3-4B-Instruct-2507 model and unified article-level benchmark protocol without changing each method's core extraction logic. For CTINexus and CTIKG, we retain their released multi-stage extraction workflows and normalize only the final outputs to our common article-level \texttt{\{entities, relations\}} JSON schema. For GraphRAG, we use the official Microsoft graph-extraction prompt and its zero-shot extraction loop. For Graphiti and Cognee, we preserve their released node/edge or cascade-extraction logic while adapting them to full-document chunked execution and article-level merging so that they can be evaluated on long CTI articles under a unified benchmark protocol. For LLM-CAKG, we keep its appendix-style prompts but adapt it to a full-document chunked processing pipeline. For AttacKG+, we retain the original rewrite-then-extract pipeline and adapt it to the same full-document article-level setting. For KnowGL, we evaluate the released HuggingFace model and parse its native decoded outputs into the same triple schema. Across prompt-based baselines, we additionally use a shared runtime wrapper for robust JSON repair and schema normalization, without changing each method's core extraction logic.

\section{Additional Training and Testing Details}
\label{app:training-details}

We collected CTI articles from public online CTI sources such as Mandiant~\cite{mandiant2024global}, CrowdStrike~\cite{crowdstrike2024threatreport}, Trend Micro~\cite{magniber2023}, Securelist~\cite{securelistthreatcategories}, Unit42~\cite{eltestcampaigns}, Microsoft Security~\cite{microsoftcyberattack}, Cisco Talos~\cite{ciscocyberattacks}, and VirusTotal~\cite{VirusTotal} to construct our CTI corpus, reserved the benchmark articles in this corpus for evaluation, and used the remaining corpus articles together with randomly sampled articles from the Primus dataset~\cite{primus} for training. For the primary Task-bank Reward setting, each training article contributes a supervision bank consisting of up to 20 four-option multi-select questions together with article-level regex targets aligned to the ground-truth KG edges. During VERL post-training, the two internal GRID extraction stages are written into a single ontology-guided extraction prompt and optimized as one generation task, rather than executed as two separate LLM calls during post-training. The 4096-token filter is applied only during training prompt construction; at test time, models receive the full article and are evaluated on full-document extraction.

Among the external sources used in evaluation, SecureNLP refers to the SemEval-2018 Task 8 shared-task dataset, an extension of MalwareTextDB~\cite{lim2017malwaretextdb,phandi2018securenlp}.

During RL training, the maximum response length is 4096 for Task-bank Reward and 8192 for End2End Reward. All experiments are run on a server with Ubuntu 20.04.6, an AMD 5955WX CPU, 256GB memory, and four Nvidia RTX 6000 Ada GPUs.

For the End2End Reward setting discussed in RQ2, we use batch size 64 and rollout count 8 for 13 RL steps. Under this configuration, the training-side online LLM-as-judge reward computation costs about \$942 in total.

\section{Reproducibility Statement}
\label{app:reproducibility}

The data artifacts, code, prompts, and evaluation scripts used in this work are available in the anonymous project repository at \url{https://github.com/anonymousauthorname/ProjectGRID}. The corresponding post-trained model weights are available in an anonymous Hugging Face repository, and the access link is also provided in the GitHub repository.

\end{document}